\documentclass[10pt,twocolumn,letterpaper]{article}

\usepackage{wacv}              % CAMERA-READY version

\usepackage[T1]{fontenc}

\usepackage{amsmath}
\usepackage{amssymb}

\usepackage{graphicx}
\usepackage{epsfig}
\usepackage{booktabs}
\usepackage{multirow}
\usepackage{colortbl}
\usepackage{float}
\usepackage{subcaption}

\usepackage[table]{xcolor}

\usepackage[
  colorlinks=true,
  linkcolor=blue,
  urlcolor=blue,
  citecolor=blue
]{hyperref}

\definecolor{wacvblue}{rgb}{0.21,0.49,0.74}

\graphicspath{{./}{./sites/}{./figures/}}

\def\wacvPaperID{*****}
\def\confName{WACV}
\def\confYear{2026}
\begin{document}

\title{Satellite-Based Detection of Looted Archaeological Sites\\ Using Machine Learning}

\author{
Girmaw Abebe Tadesse$^{1}$\thanks{Corresponding author: \texttt{gtadesse@microsoft.com}} \quad
Titien Bartette$^{2}$ \quad
Andrew Hassanali$^{3}$ \quad
Allen Kim$^{1}$ \quad
Jonathan Chemla$^{2}$ \quad \\
Andrew Zolli$^{3}$ \quad
Yves Ubelmann$^{2}$ \quad
Caleb Robinson$^{1}$ \quad
Inbal Becker-Reshef$^{1}$ \quad
Juan Lavista Ferres$^{1}$ \\
\\
$^{1}$Microsoft AI for Good Research Lab\\
$^{2}$Iconem\\
$^{3}$Planet Labs PBC
}

\maketitle
	\thispagestyle{empty}

\begin{abstract}
Looting at archaeological sites poses a severe risk to cultural heritage, yet monitoring thousands of remote locations remains operationally difficult. We present a scalable and satellite-based pipeline to detect looted archaeological sites, using PlanetScope monthly mosaics (4.7m/pixel) and a curated dataset of 1,943 archaeological sites in Afghanistan (898 looted, 1,045 preserved) with multi-year imagery (2016--2023) and site-footprint masks. We compare (i) end-to-end CNN classifiers trained on raw RGB patches and (ii) traditional machine learning (ML) trained on handcrafted spectral/texture features and embeddings from recent remote-sensing foundation models. Results indicate that ImageNet-pretrained CNNs combined with spatial masking reach an F1 score of $0.926$, clearly surpassing the strongest traditional ML setup, which attains an F1 score of $0.710$ using SatCLIP-V+RF+Mean, i.e., location and vision embeddings fed into a Random Forest with mean-based temporal aggregation. Ablation studies demonstrate that ImageNet pretraining (even in the presence of domain shift) and spatial masking enhance performance. In contrast, geospatial foundation model embeddings perform competitively with handcrafted features, suggesting that looting signatures are extremely localized. The repository is available at \url{https://github.com/microsoft/looted_site_detection}.
\end{abstract}

\section{Introduction}
The looting of archeological sites represents one of the most critical threats to global cultural heritage. With a growing threat of archeological sites around the world, many located in remote or conflict-affected regions, such as the Middle East and North Africa~\cite{casana2017satellite,brodie2018illegal}, traditional ground-based and manual monitoring cannot scale effectively, which are labor-intensive, costly, and often infeasible for large or remote regions~\cite{parcak2016satellite}. 
The growing availability of high-resolution, high-frequency satellite imagery has enabled remote sensing to become a powerful tool for archaeological site monitoring~\cite{parcak2016satellite-evidence}.  A recent survey assesses advances and practical challenges for heritage monitoring across Europe~\cite{cuca2023monitoring}. Policy and community guidance has emphasized Earth Observation (EO) to safeguard cultural and natural heritage~\cite{unesco2015_eo_heritage, levin2019world}.

Satellite-based remote sensing data have long been used to aid in identifying looting activities at archaeological sites~\cite{menze2012,tapete2018detection}. Traditionally, these analyses have depended on manual examination, occasionally supported by basic image processing methods, such as edge enhancement. However, such manual workflows are subjective and not scalable for monitoring large areas or long temporal sequences. in addition, archaeological looting produces subtle and spatially diffuse surface alterations, including disturbed soil texture, irregular micro-relief, and localized spectral anomalies. These signals are often visually ambiguous and can overlap with other processes such as erosion or agricultural activity, making automated discrimination particularly challenging. Recent advances in machine learning offer a more scalable and consistent alternative, enabling the automatic detection of looted sites~\cite{vincent2025detecting}. 

Recent studies reveal a number of foundation models for remote sensing, such as SatMAE~\cite{cong2022satmae}, SatCLIP~\cite{klemmer2024satclip}, DINOv3~\cite{siméoni2025dinov3}, and Prithvi-EO~\cite{jakubik2023foundation}, where their embeddings have been demonstrated to be effective across a wide range of downstream tasks.
However, the effectiveness of these embeddings for archaeological monitoring remains unexplored. While handcrafted features, such as spectral signals from the Near-Infrared (NIR) band and texture features, can be extracted to highlight looting-related patterns, they have not been systematically evaluated in state-of-the-art pipelines, nor compared against embeddings extracted from these foundation models. Similarly, the effectiveness of conventional machine-learning classifiers, such as Random Forest, in detecting looted sites, and how they compare to CNN-based methods that leverage transfer learning through ImageNet pretraining~\cite{yosinski2014transferable}, has not yet been investigated.

\begin{figure}[t]
    \centering
    \includegraphics[width=0.45\textwidth]{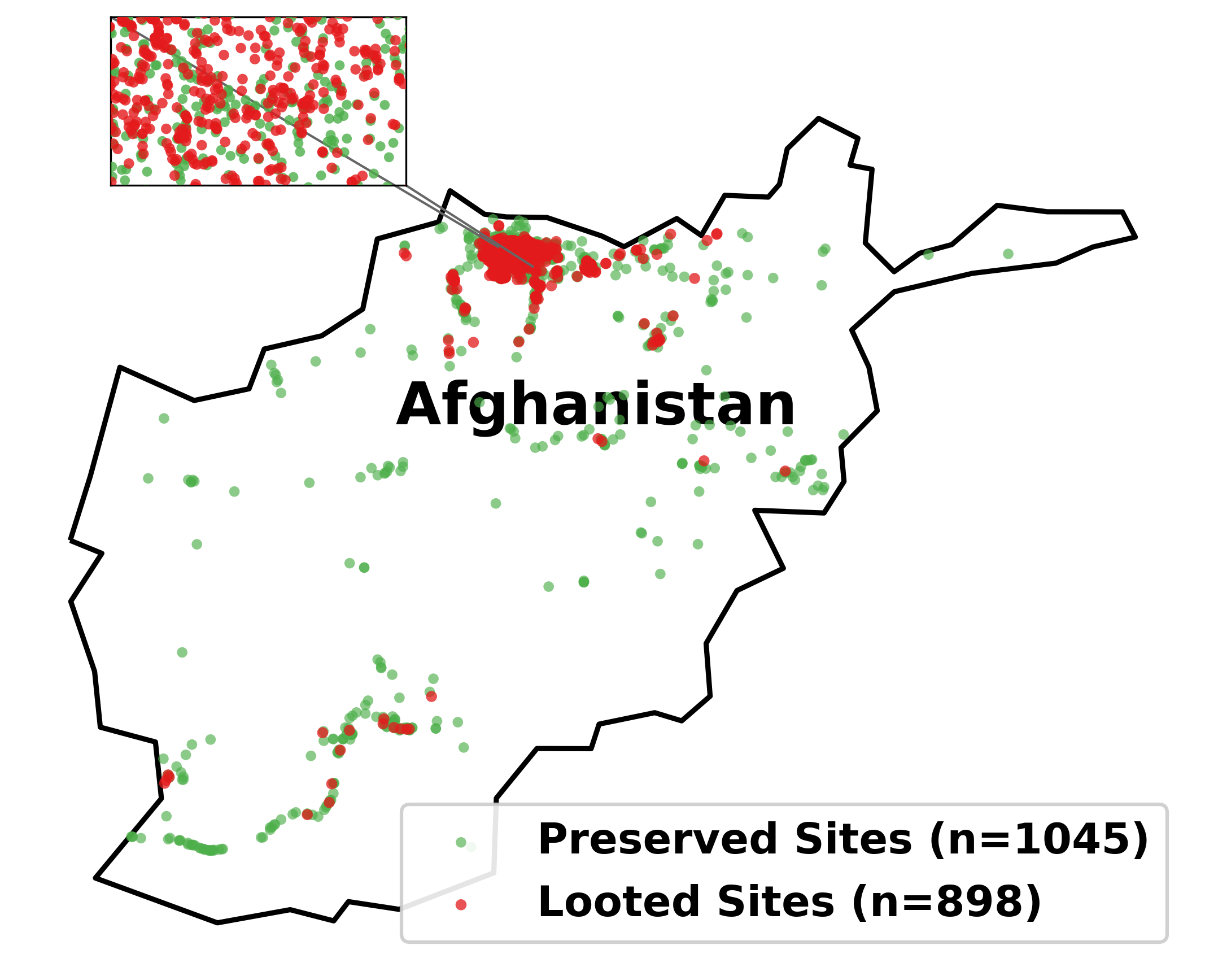}
    \includegraphics[width=0.45\textwidth]{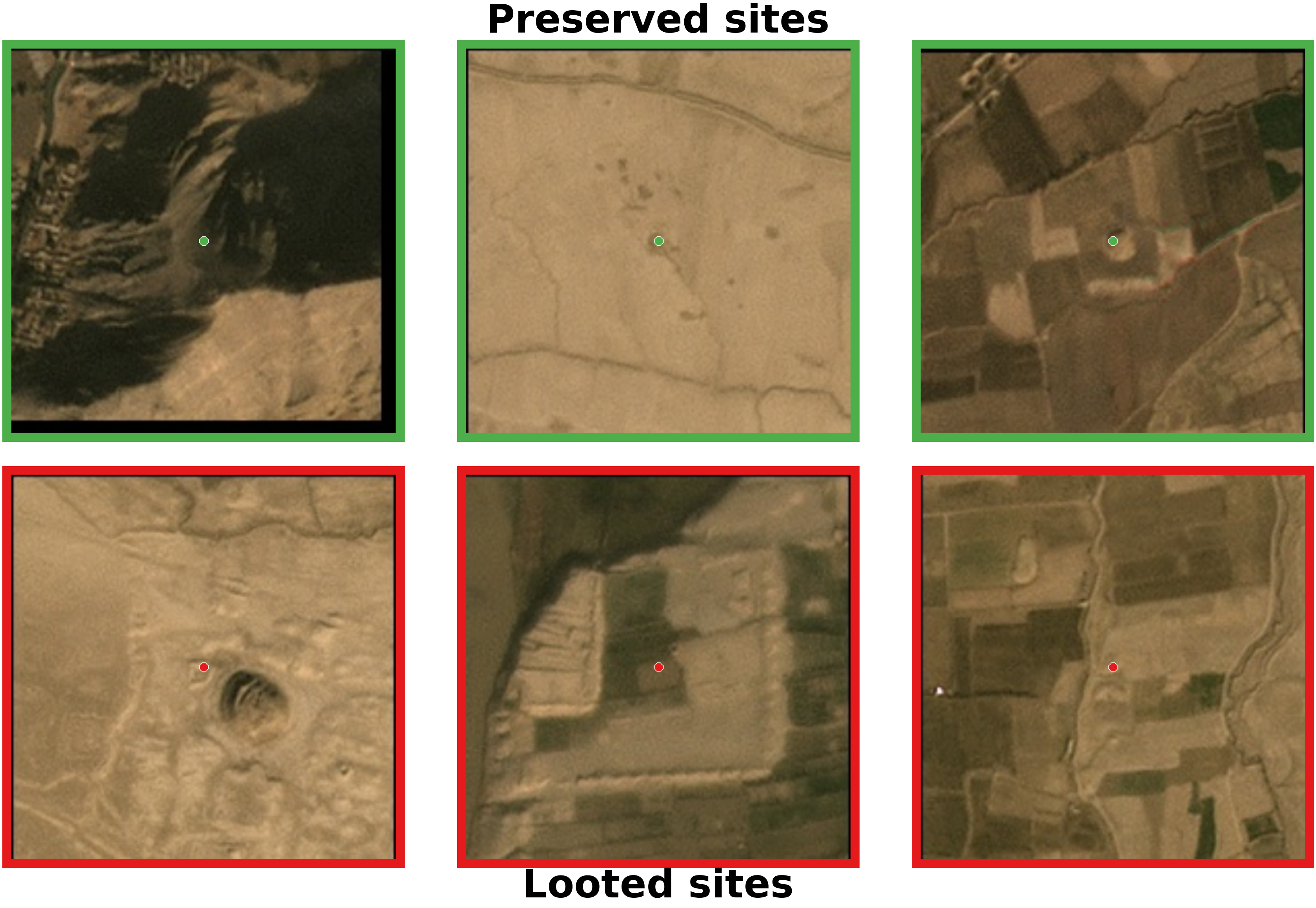}
    \caption{Overview of the archaeological sites in Afghanistan in this work. The sites are composed of $1045$ \textcolor{green}{preserved} and $898$ \textcolor{red}{looted} sites.}
    \label{fig:sites}
\end{figure}

In this work, we propose a satellite-based approach for detecting looted archaeological sites in Afghanistan (see Figure~\ref{fig:sites}), using CNN-based deep learning models and PlanetScope monthly mosaics (4.7 m/pixel resolution).
Our main contributions are as follows: \textbf{1.~Comprehensive Framework:} We present the first systematic comparison between CNN-based models (trained directly on raw imagery) and traditional ML methods for archaeological looting detection, evaluating five CNN architectures and seven feature families (one handcrafted feature set and six embedding sets from foundation models).
 \textbf{2.~Temporal Consistency Insight:} We show that training exclusively on a single year imagery achieves higher accuracy than multi-year training by mitigating temporal label noise in looted-site annotations and cumulative process of looting activities.
\textbf{3.~Quantified Transfer Learning and Spatial Masking Benefits:} We demonstrate that ImageNet pretraining provides a $6-14\%$ F1 improvement, supported by detailed architectural ablations across CNN backbones. Manually-annotated spatial masking results in $30 - 45\%$ F1 improvements.
\textbf{4.~Large-Scale Dataset Contribution:} We assemble the largest dataset of archaeological sites ( $1,943$), with multi-year (2016-2023) imagery and manually-annotated spatial masks. 

% We focus on three questions: (1) Do CNNs trained on raw imagery outperform feature-based ML for looting detection? (2) How much do transfer learning and spatial masking matter? (3) How robust are models to temporal variation across years?

\begin{table*}[ht]
    \centering
    \resizebox{0.9\textwidth}{!}{
    \begin{tabular}{c c c c c c c c c c}
        &\textbf{2016} & \textbf{2017} & \textbf{2018} & \textbf{2019} & \textbf{2020} & \textbf{2021} & \textbf{2022} & \textbf{2023} & \textbf{Mask} \\
        \multirow{2}{*}{Preserved}&\includegraphics[width=0.1\textwidth]{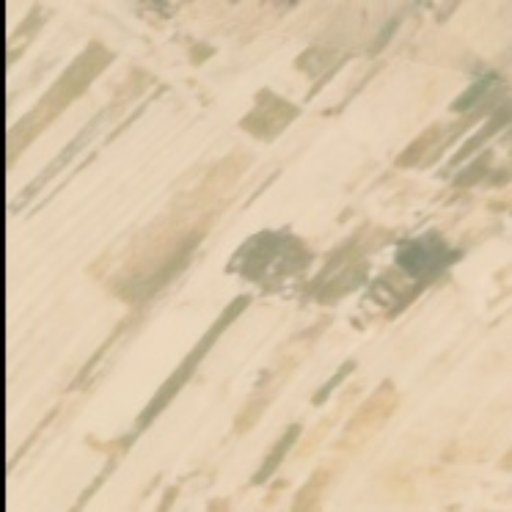} &
        \includegraphics[width=0.1\textwidth]{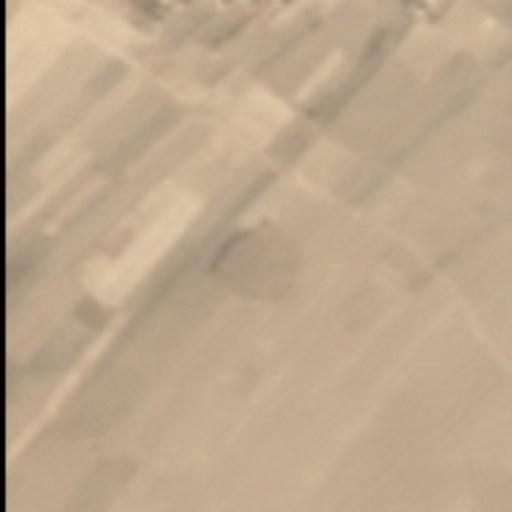} &
        \includegraphics[width=0.1\textwidth]{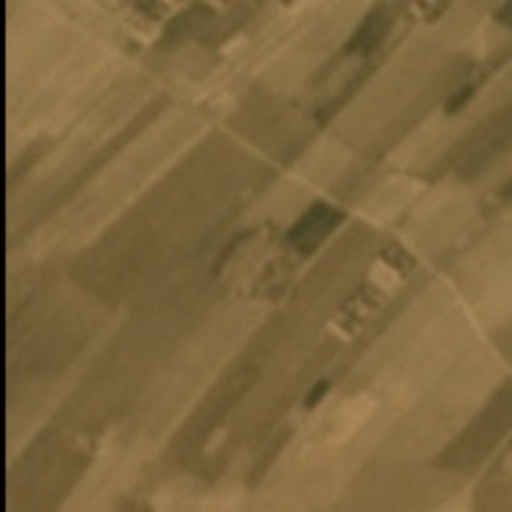} &
        \includegraphics[width=0.1\textwidth]{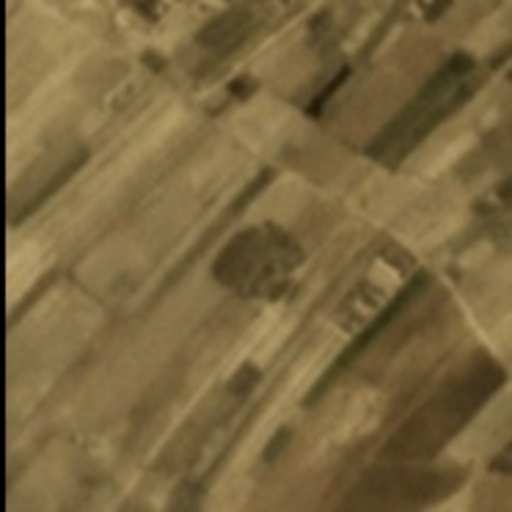} &
        \includegraphics[width=0.1\textwidth]{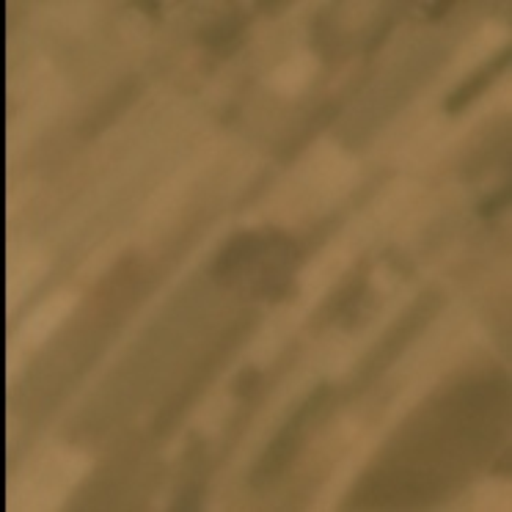} &
        \includegraphics[width=0.1\textwidth]{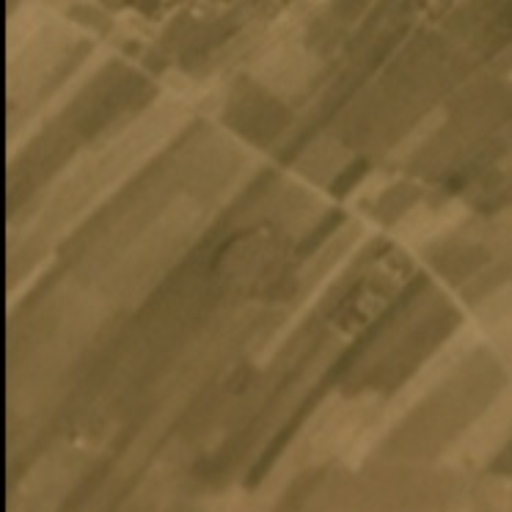} &
        \includegraphics[width=0.1\textwidth]{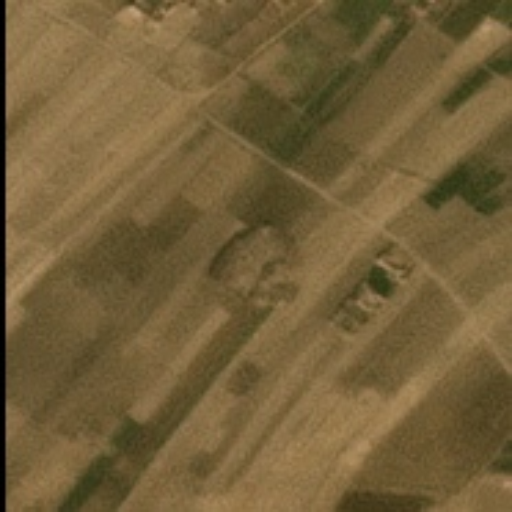} &
        \includegraphics[width=0.1\textwidth]{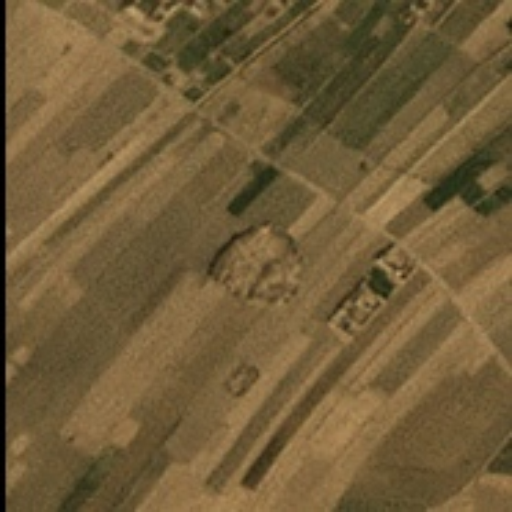} &
        \includegraphics[width=0.1\textwidth]{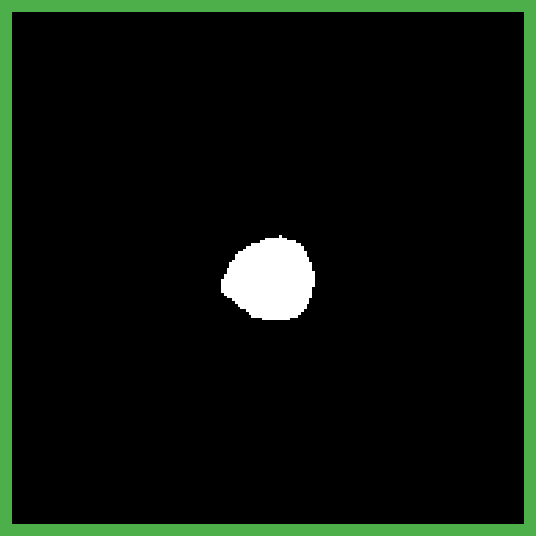} \\ 
        &\includegraphics[width=0.1\textwidth]{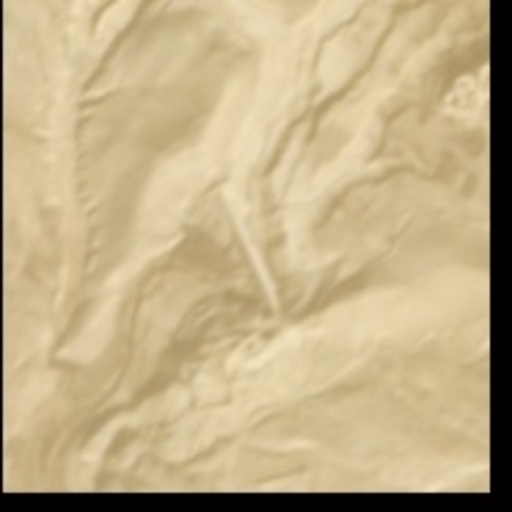} &
        \includegraphics[width=0.1\textwidth]{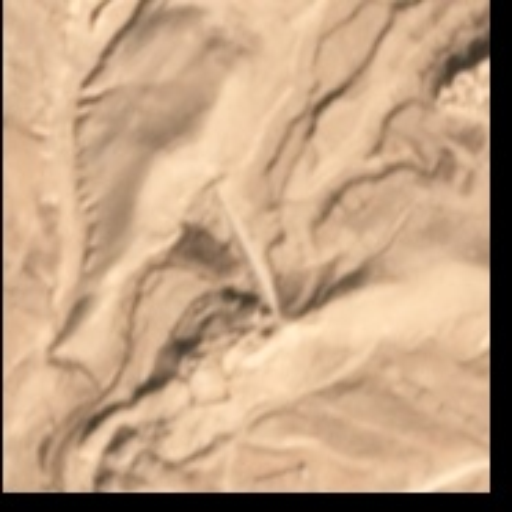} &
        \includegraphics[width=0.1\textwidth]{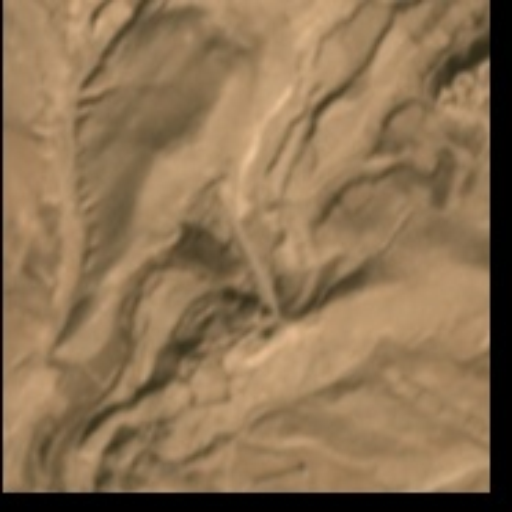} &
        \includegraphics[width=0.1\textwidth]{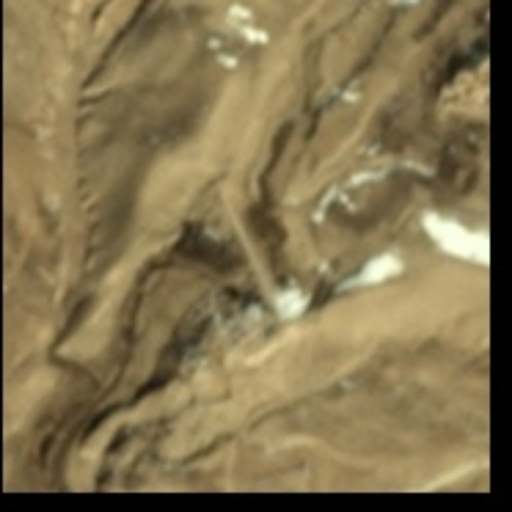} &
        \includegraphics[width=0.1\textwidth]{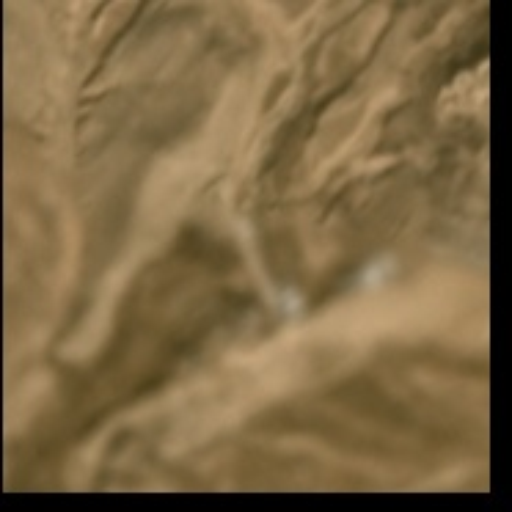} &
        \includegraphics[width=0.1\textwidth]{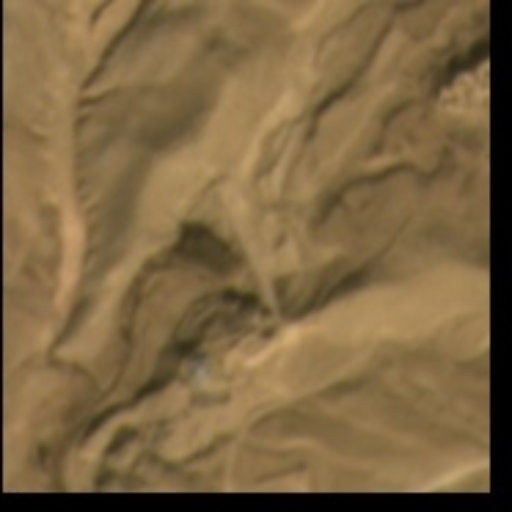} &
        \includegraphics[width=0.1\textwidth]{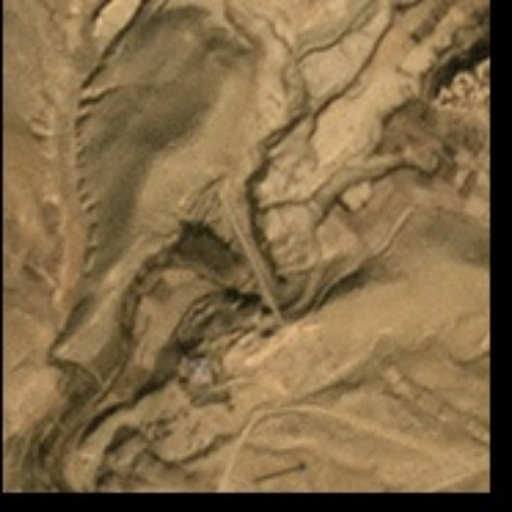} &
        \includegraphics[width=0.1\textwidth]{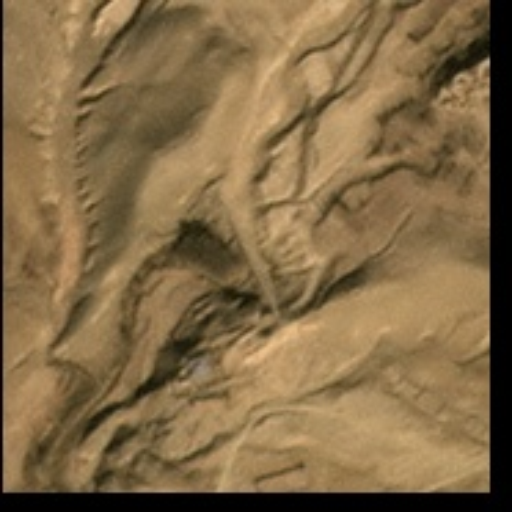} &
        \includegraphics[width=0.1\textwidth]{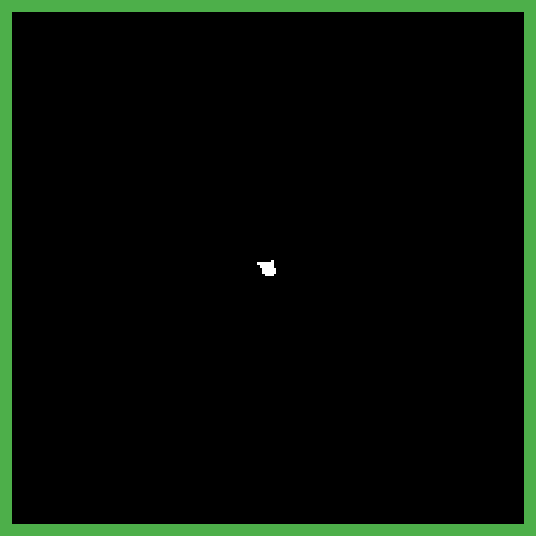} \\ \hline

           \multirow{2}{*}{Looted} &\includegraphics[width=0.1\textwidth]{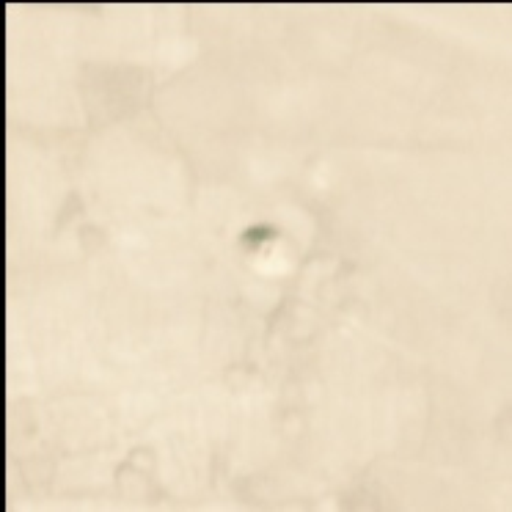} &
        \includegraphics[width=0.1\textwidth]{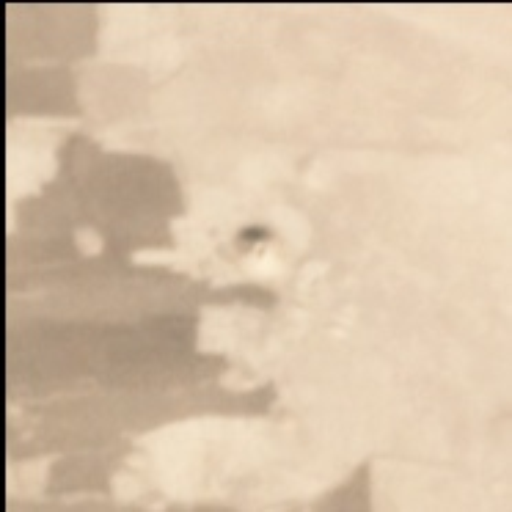} &
        \includegraphics[width=0.1\textwidth]{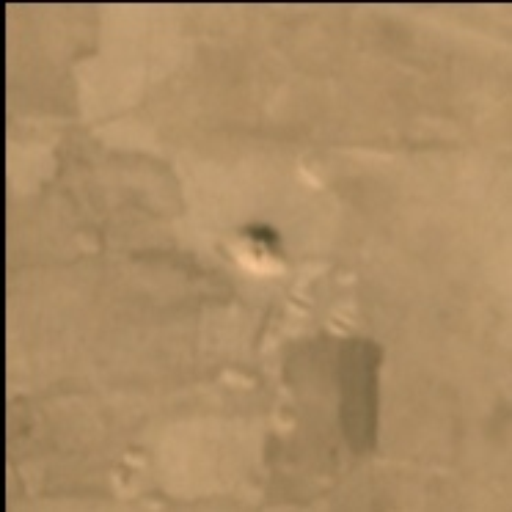} &
        \includegraphics[width=0.1\textwidth]{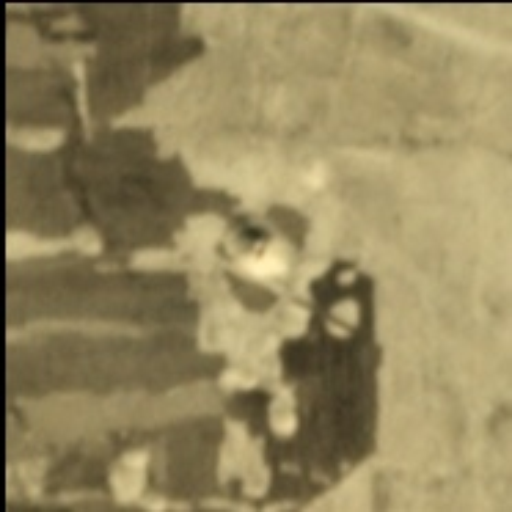} &
        \includegraphics[width=0.1\textwidth]{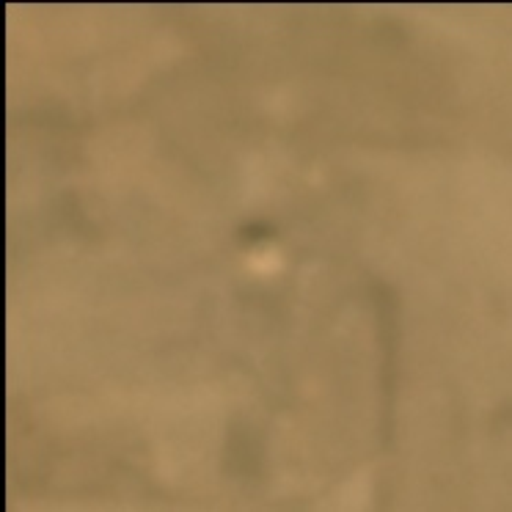} &
        \includegraphics[width=0.1\textwidth]{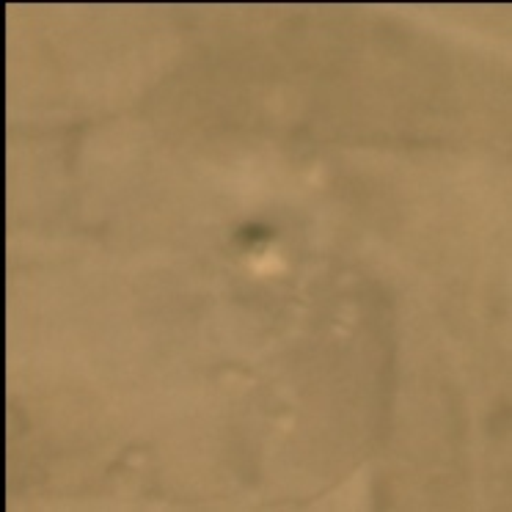} &
        \includegraphics[width=0.1\textwidth]{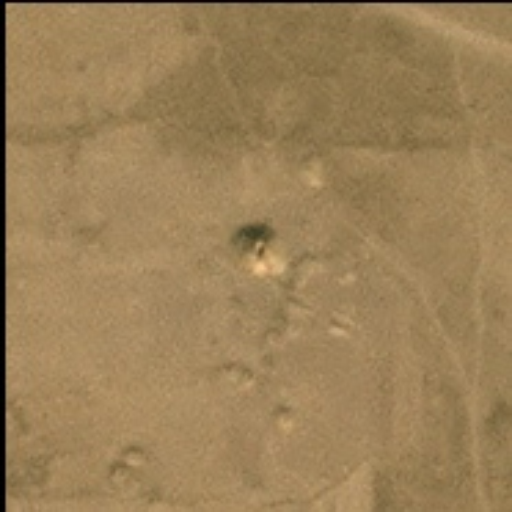} &
        \includegraphics[width=0.1\textwidth]{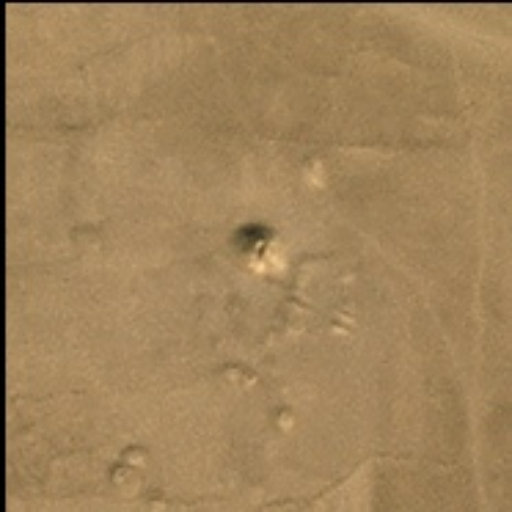} &
        \includegraphics[width=0.1\textwidth]{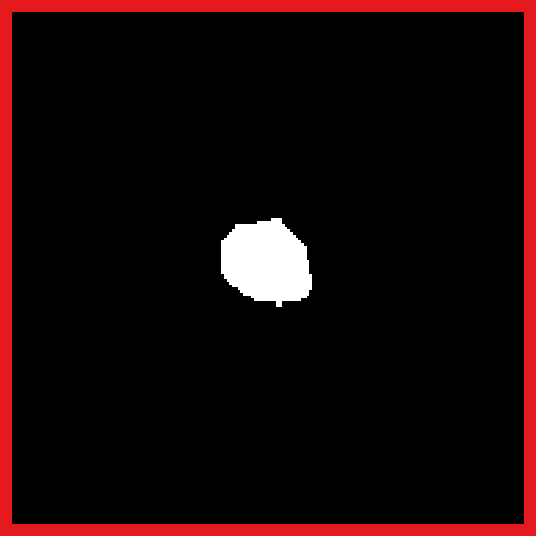} \\ 
       & \includegraphics[width=0.1\textwidth]{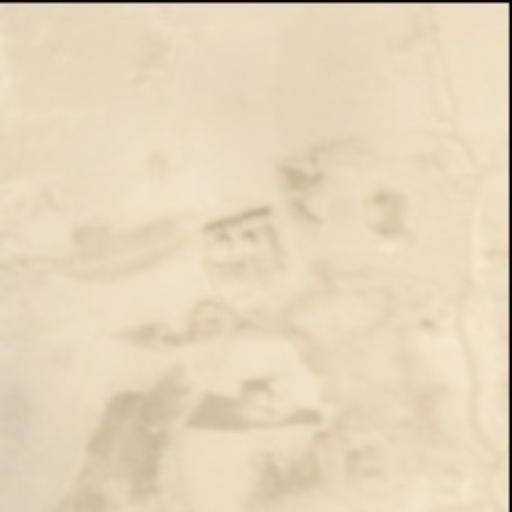} &
        \includegraphics[width=0.1\textwidth]{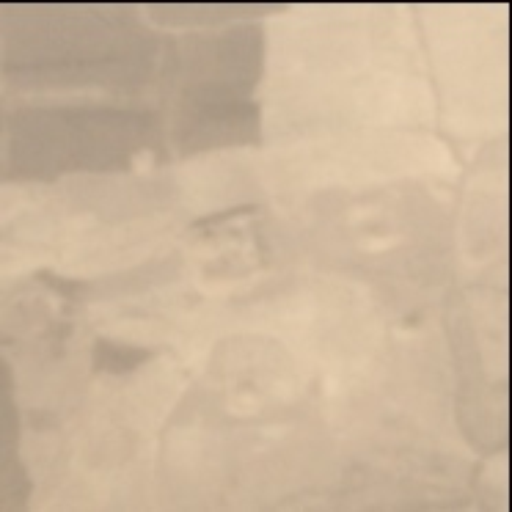} &
        \includegraphics[width=0.1\textwidth]{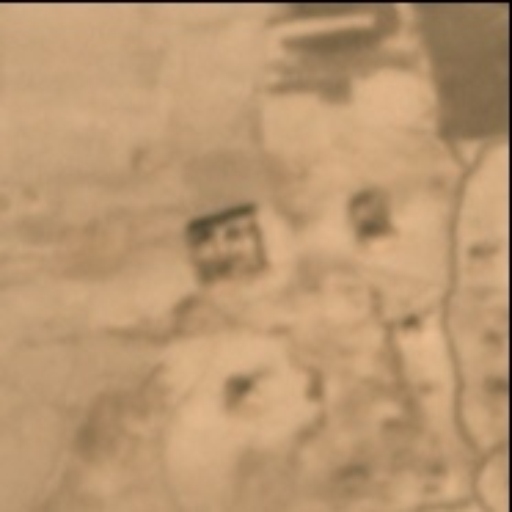} &
        \includegraphics[width=0.1\textwidth]{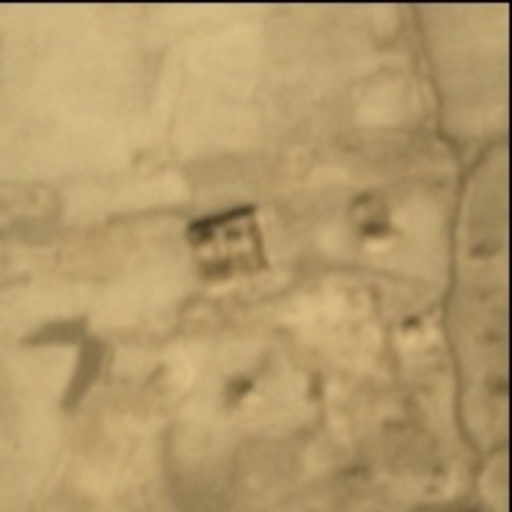} &
        \includegraphics[width=0.1\textwidth]{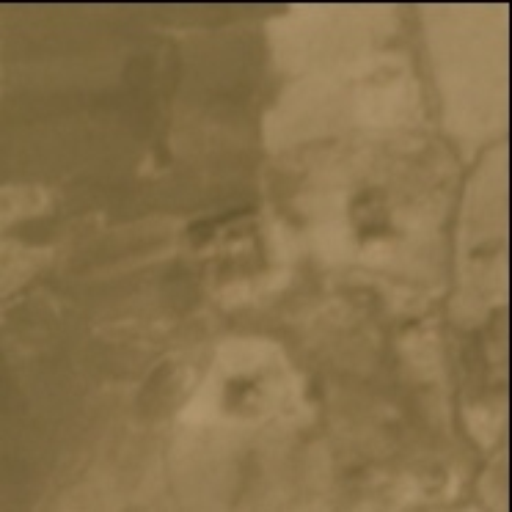} &
        \includegraphics[width=0.1\textwidth]{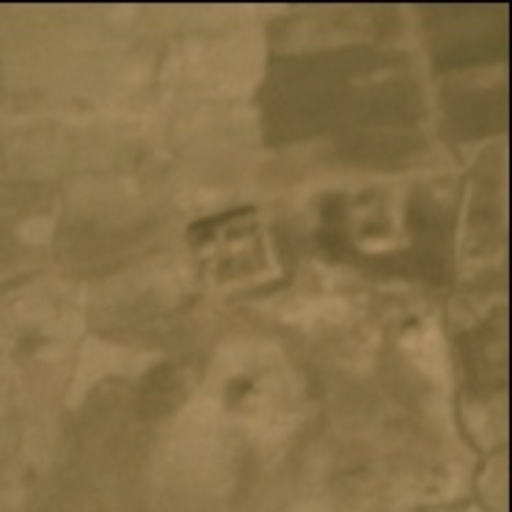} &
        \includegraphics[width=0.1\textwidth]{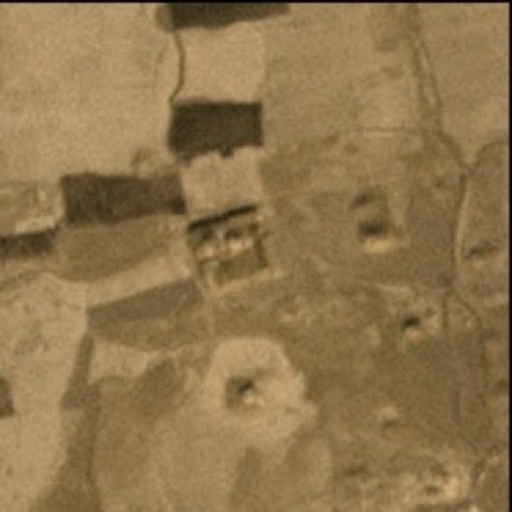} &
        \includegraphics[width=0.1\textwidth]{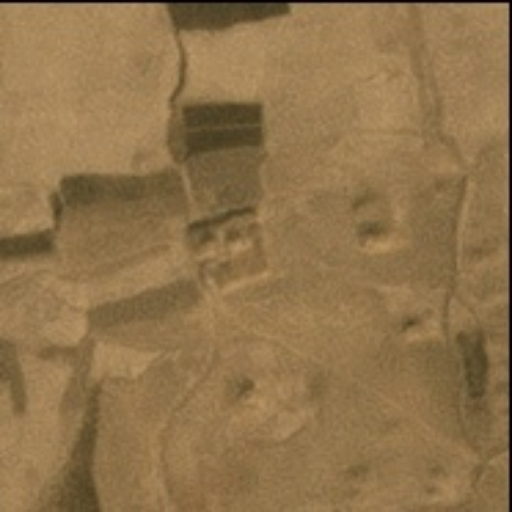} &
        \includegraphics[width=0.1\textwidth]{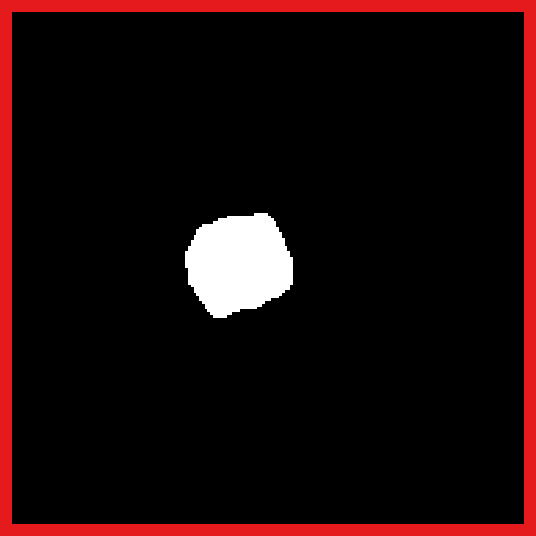} \\  
    \end{tabular}
    }
    \caption{PlanetScope monthly mosaics and their corresponding binary masks for selected locations, shown for December in each year from 2016 through 2023.The (latitude, longitude) coordinates for each site were supplied by our partners at \href{https://iconem.com/}{Iconem}, and based on these coordinates, the 1 km × 1 km monthly mosaic was downloaded for each site. The masks were annotated manually. Looted sites exhibit disturbed soil tone/texture; preserved sites show uniform surfaces.}\label{fig:qual_examples}
\end{table*}

\section{Related Work}
\textbf{Detection of archaeological looting.} Early work relied on manual visual interpretation of multispectral imagery~\cite{parcak2016satellite}. Subsequent studies demonstrated large-scale monitoring through temporal high-resolution imagery and systematic documentation of looting density~\cite{casana2017satellite}. Classical image processing and change detection methods (e.g., differencing, PCA) offer modest improvements but still require manual feature engineering and are difficult to scale robustly~\cite{tapete2018detection}. Recent efforts explore CNNs for related pit detection problems but are limited by small labeled datasets~\cite{caspari2019convolutional}. In contrast, we evaluate looting detection at larger scale (1,943 sites) and provide spatial footprint masks as an intrinsic attention mechanism. 

\textbf{CNNs and transfer learning for satellite imagery.} Transfer learning from ImageNet has proven effective for a wide range of tasks, even when there is a domain shift between source and target data~\cite{yosinski2014transferable}. The ResNet~\cite{he2016resnet} and EfficientNet~\cite{tan2019efficientnet} model families continue to serve as strong baselines for image classification; however, systematic comparisons with traditional ML approaches for archaeological looting detection, and the roles of spatial masking and pretraining for looting detection, remain scarce.

\textbf{Foundation models for remote sensing and temporal analysis.} Recent remote-sensing foundation models (e.g., SatMAE~\cite{cong2022satmae}, Prithvi-EO~\cite{jakubik2023foundation,szwarcman2024prithvi}, SatCLIP~\cite{klemmer2024satclip}, GeoRSCLIP~\cite{zhang2024georsclip}, and DINOv3~\cite{siméoni2025dinov3} provide general-purpose representations, yet their utility for subtle archaeological disturbance is not well established. In addition, comparisons of geospatial embeddings with handcrafted spectral statistics and texture features for looting pattern detection remain unexplored.

\textbf{Aggregation of temporal features.} For time series satellite imagery like our multi-year monthly mosaics, sequence-based models, for example temporal self-attention frameworks~\cite{garnot2020lightweight}, can perform well but encounter difficulties when handling very long sequences, such as our dataset spanning monthly mosaics from 2016 to 2023. Alternatives include statistical pooling and PCA-based compression~\cite{pelletier2016temporal,russwurm2018temporal,rodarmel2002pca}. We compare multiple aggregation strategies such as mean, median and concatenation in addition to PCA. We also analyze the temporal consistency of the labels.

\textbf{Datasets of archaeological sites.}
Existing aerial and satellite datasets for detecting looted archaeological site differ markedly in terms of accessibility, spatial resolution, and geographic as well as temporal coverage. Masini et al.~\cite{masini2019recent} provide multi-temporal satellite imagery for two Syrian sites, with varying spatial resolutions and annual updates. El Hajj~\cite{el2021interferometric} provides high-resolution imagery ($15 m/pixel$) for nine locations in Syria and Iraq, though these areas are not imaged on a regular basis. Payntar~\cite{payntar2023multi} assembles a Peruvian dataset at $30 m/pixel$ resolution that includes $477$ sites, each captured once over a five-year span. Altaweel et al.~\cite{altaweel2024monitoring} utilize UAV imagery at $3 cm/pixel$ resolution to document $95$ sites worldwide, but this collection similarly does not include repeated temporal coverage. Vincent et al.~\cite{vincent2025detecting} present a multi-temporal dataset consisting of monthly satellite images spanning $675$ sites in northern Afghanistan. Our dataset includes the greatest number of sites to date (roughly $3 \times$ more than the previous maximum reported in~\cite{vincent2025detecting}) and incorporates looted locations from across the entire Afghan region (see Figure~\ref{fig:sites}).

%%%%%%%%%%%%%%%%%%%%%%%%%%%%%%%%%%%%%%%
\section{Dataset}
\textbf{Sites and labels.} Our dataset contains 1,943 archaeological sites in Afghanistan: 898 looted (46.2\%) and 1,045 preserved (53.8\%), collected from \textit{Archaeological gazetteer of Afghanistan}~\cite{ball2019archaeological}  and further verified by a team of expert archaeologists, who used online platforms with high-resolution satellite imagery (Google Earth, ESRI, and Bing) to expand the existing list with additional sites and classify them into the ‘looted’ or ‘preserved’ categories, depending on whether they had been affected by deliberate human damage before 2023 or not (see Table~\ref{tab:dataset_stats}).

        	\textbf{PlanetScope imagery.} We use PlanetScope monthly mosaics (RGB+NIR) at 4.7m/pixel spanning January 2016--December 2023 (96 months). For each site and month we construct a median composite mosaic and extract a site-centered area of interest (approximately 1km$\times$1km). See Table~\ref{fig:qual_examples} for some examples of looted and preserved sites over the years.

        	\textbf{Spatial masks.} We manually annotate polygon site footprints and rasterize them into binary masks aligned to each patch. Masks suppress non-site context such as roads, modern settlements, and agricultural fields. Note that footprints are archaeologically informed estimates of the wider site boundaries, covering both visible surface remains and nearby zones where related buried or peripheral activities are likely to occur. Consequently, the masks capture a managed degree of spatial uncertainty rather than an exact ground-truth segmentation. Thus, the spatial masks for looted sites are often larger than those for preserved sites, because looting can occur both within the site boundaries and in the surrounding area.

        	\textbf{Temporal label consistency.} Labels are assigned as of December 2023 (looted vs. preserved). When training on imagery far earlier than the label timestamp, a subset of sites labeled \emph{looted} may visually appear intact in earlier years, introducing temporal label noise. This decision acknowledges that looting typically unfolds as a cumulative process, leaving visible traces that can remain for extended periods, even after active digging has stopped. In this work, we aim to detect sites in a degraded condition at the state level, rather than to pinpoint the exact timing of individual looting episodes. We therefore emphasize temporally consistent 2023 imagery for primary comparisons while separately evaluating across-year robustness.

\begin{table}[t]
\centering
\caption{Dataset statistics summary.}
\label{tab:dataset_stats}
\small
\rowcolors{2}{gray!12}{white}
\begin{tabular}{@{}lr@{}}
        \toprule
        \textbf{Attribute} & \textbf{Value} \\
\midrule
Total Sites & 1,943 \\
Looted Sites & 898 (46.2\%) \\
Preserved Sites & 1,045 (53.8\%) \\
Geographic Coverage & Afghanistan \\
Temporal Span & 2016--2023 (96 months) \\
Spatial Resolution & 4.7m/pixel \\
Patch Size & $186\times186$ \\
\bottomrule
\end{tabular}
\end{table}

\section{Methods}
We evaluate two methodological families: (1) CNNs trained end-to-end on raw RGB imagery, and (2) traditional ML classifiers trained on temporally aggregated features (handcrafted spectral/texture features and embeddings from geospatial foundation models). In this Section, we present the handcrafted features, embeddings, models, and types of temporal aggregation used in this work.

\subsection{Handcrafted features}
We extract a set of manually designed features that capture spectral properties and textural patterns from PlanetScope monthly mosaics (RGB + NIR), designed to capture both abrupt and gradual changes indicative of mechanical and manual looting activities. 
The proposed diverse feature set jointly captures spectral anomalies from soil/vegetation disturbance, spatial irregularities from excavation patterns, and temporal change dynamics. Moreover, standard deviation, coefficient of variation, and temporal slope over multi-month windows detect gradual degradation.

\paragraph{Feature examples.}
Given a site-masked monthly mosaic with red ($R$), green ($G$), blue ($B$), and near-infrared ($NIR$) bands, we extract spectral statistics such as vegetation and moisture indices as well as extra disturbance-sensitive indices and summary statistics (means, standard deviations) from each band.  We also include texture features such as Gray-Level Co-occurrence Matrix (GLCM) that quantify spatial structure indicative of disturbance statistics (i.e., contrast, homogeneity, energy and entropy) and Local Binary Pattern (LBP) that encodes micro-texture via local thresholding on a circular neighborhood. Once each band is quantized into a set of discrete levels, increased contrast and entropy, together with reduced homogeneity and energy, indicate disturbed or looted patterns. Overall, we computed 42-dimensional handcrafted features (12 features and 30 texture features) per monthly composite capturing vegetation health, soil properties, and texture.

% \begin{table}[t]
%     \centering
%     \small
%     \caption{Foundation models used for embedding extraction, with the backbone family, approximate number of parameters and dimensions.}
%     \label{tab:foundation_features}
%     \rowcolors{2}{gray!12}{white}
%     \setlength{\tabcolsep}{6pt}
%     \begin{tabular}{l l r r}
%         \toprule
%         \textbf{Feature} & \textbf{Backbone} & \textbf{Params (M)} & \textbf{Dim} \\
%         \midrule
%         DINOv3 & ViT-L/16 & 304 & 1024 \\
%         GeoRSCLIP & ViT-B/32 & 86 & 512 \\
%         Prithvi-EO 2.0 & ViT-H/14 & 600 & 1024 \\
%         SatCLIP & ViT-B/16 & 86 & 512 \\
%         Satlas Pretrain & ResNet-152 & 60 & 2048 \\
%         SatMAE & ViT-B/8 & 86 & 768 \\
%         \bottomrule
%     \end{tabular}
% \end{table}
\subsection{Embeddings from foundation models}
We extract embeddings from monthly Planet mosaics using geospatial foundation models. Each backbone applies its native preprocessing protocol and produces embeddings at its canonical dimensionality. These foundational models and their corresponding embeddings are presented as follows.
\textbf{DINOv3:} We use the ViT-L/16 image encoder pre-trained on the SAT‑493M dataset~\cite{siméoni2025dinov3}. Inputs are 3‑channel RGB with SAT‑493M normalization, and we extract 1024‑dimensional embedding.
\textbf{GeoRSCLIP:} We employ OpenCLIP's ViT-B/32 backbone with RS5M-trained weights~\cite{zhang2024georsclip}. The CLIP image encoder produces 512-dimensional embeddings following standard CLIP preprocessing.
\textbf{Prithvi-EO 2.0:} We instantiate the 600M-parameter ViT-Huge/14 encoder~\cite{szwarcman2024prithvi} 
% using the authors' \texttt{PrithviViT} implementation with
with \texttt{embed\_dim=1280}, \texttt{depth=32}, and \texttt{patch\_size=[1,14,14]}. The model is adapted for single-frame Planet 4-band input, with CLS tokens projected to 1024 dimensions via a learned linear layer.
\textbf{SatCLIP:} We use the location encoder from~\cite{klemmer2024satclip} via Hugging Face to obtain 256-dimensional embeddings for each site. We then concatenate these with 512-dimensional embeddings extracted from the pooled image encoder output of an RGB-based CLIP vision model (ViT-B/16) from each monthly imagery, resulting in a unified 768-dimensional embedding, which we hereafter refer to as SatCLIP-V.
% The image processor is constructed from \texttt{vision\_config} when unavailable.
%
\textbf{Satlas-Pretrain:} We employ ResNet-152~\cite{bastani2023satlaspretrain} with the Sentinel-2 multi-image/multi-scale checkpoint, removing the classification head to expose global average pooled features, yielding 2048-dimensional embeddings. 
% The first convolutional layer is adapted for 9-channel input when required, yielding 2048-dimensional embeddings.
%
\textbf{SatMAE:} We instantiate a group-channel masked autoencoder~\cite{cong2022satmae} with a ViT-Base encoder (\texttt{embed\_dim=768}, \texttt{depth=12}, \texttt{patch\_size=8}),  adapted to a single 4-band group for Planet imagery. Preprocessing applies per-band min-max normalization followed by standardization, yielding 768-dimensional embeddings.

\subsection{Models and Temporal Aggregation}

\paragraph{Traditional ML models on features:}
We evaluate Logistic Regression (LR), Random Forest (RF), Gradient Boosting (GB), and XGBoost (XGB) that are trained two feature groups: handcrafted and embeddings.
 Given monthly features $\{\mathbf{x}_t\}_{t=1}^{T}$, we compare different temporal aggregation approaches: Mean, Median, Concatenation, and PCA-based compression of monthly features.

\paragraph{CNN models on raw RGB patches:}
We employed two families of CNN architectures to learn looting patterns directly from raw imagery: ResNet~\cite{he2016resnet} and EfficientNet~\cite{tan2019efficientnet}.
ResNet models employ identity skip connections to stabilize training of deeper networks and achieve strong classification accuracy on natural images; we span depths (18/34/50) to probe capacity effectst~\cite{he2016resnet}. EfficientNet uses compound scaling to balance accuracy and efficiency; B0/B1 provide competitive performance under constrained compute and memory, making them suitable for operational monitoring pipelines~\cite{tan2019efficientnet}.
We evaluate ResNet-18/34/50 and EfficientNet-B0/B1 with ImageNet initialization or random initialization.
We train these models both with and without spatial masking. When masking is used, we apply a binary mask $M\in\{0,1\}^{224\times224}$ to the RGB patch $I$ at each site via element-wise multiplication, yielding $I_{\text{masked}} = I\odot M$.

% To force this table to the top of the Results page/section, place it
% right after the \section{Results} line in your LaTeX source,
% and use the [t] placement option with the starred table environment as done above.
% If using a two-column class (e.g., IEEEtran), also compile twice and
% consider the `stfloats` or `fixltx2e` packages if float placement is stubborn.

\section{Experimental Setup}

\paragraph{Preprocessing.}
Handcrafted features and embeddings are extracted from the raster images. The images were resized to $224 \times 224$ for embedding extraction. When masking is used, features and embeddings are computed exclusively from regions where the mask has non-zero pixel values. Polygon footprints are converted into raster masks at the mosaic resolution, and both the mosaics and the masks are aligned to be centered at the latitude and longitude of each site. The raster images are then converted to RGB images for the CNN models. Images from 2016 were excluded from the experiments because of quality issue. 

\paragraph{ Data splits and metrics.} We apply stratified splitting at the site level ensuring that each site appears in only one of the train, validation and test splits. The default split is 70\% train (1,359 sites), 10\% validation (195), 20\% test (389), stratified by class. We report 5-fold cross-validation.
% (fold seeds $42+k$).
%
We report accuracy, precision, recall, F1 score (positive class =looted), and AUROC on 389 test sites (20\%) for each fold, with mean $\pm$ std across the five folds. 
\paragraph{Augmentation and training details.}
For CNNs we apply horizontal/vertical flips (p=0.5), small rotations (\(\pm 10^\circ\)), random brightness/contrast jitter, and slight Gaussian noise. We train with Adam (lr $3\times 10^{-4}$, weight decay $0.01$), StepLR (step=20, $\gamma=0.1$), batch size 32, up to 60 epochs, selecting the best epoch by validation F1 with early stopping (patience 10). Class imbalance is handled by stratified folds and loss weighting by inverse class frequency.

% \paragraph{Model selection and reporting.}
% All metrics are computed per-fold on the held-out test split at the default threshold 0.5 and then aggregated as mean$\pm$std across folds. AUROC is threshold-free and computed from raw logits. We verify seeds for dataloader shuffling and model initialization to ensure fold reproducibility.

\paragraph{Traditional ML configuration.}
Handcrafted and foundation-model features are computed monthly, aggregated by the different strategies: mean, median, concatenation and PCA. PCA is fit only on training folds,  the dimensionality is reduced to $1024$ whenever the original feature dimension exceeds the number of training samples, e.g., in case of Satlas-pretrain embeddings. We tune LR (C), RF (trees, depth), GB and XGB (learning rate, estimators, depth) via 3-fold CV on the training split.

\section{Results}
\begin{table*}[htbp]
\centering
\caption{Performance of CNN and conventional models on 2023 imagery. Values are mean $\pm$ std across 5 folds. Larger numbers are better for all metrics. The Traditional ML results correspond to the best-performing combination of classifier and temporal aggregation strategy for each feature type. Abbreviations: XGB = XGBoost, LR = Logistic Regression, GB = Gradient Boosting, RF = Random Forest. \textbf{Bolded} values indicate the highest mean scored for each metric within a model group.}
\label{tab:cnn_and_traditional_results}
\small
\rowcolors{2}{gray!12}{white}
\setlength{\tabcolsep}{5pt}
\resizebox{0.85\textwidth}{!}{%
\begin{tabular}{l c c c c c}
\toprule
\textbf{Model / Configuration} & \textbf{Accuracy} & \textbf{Precision} & \textbf{Recall} & \textbf{F1} & \textbf{AUROC} \\
\midrule
\multicolumn{6}{l}{\textbf{CNN Models}} \\
\midrule
EfficientNet-B0 & 0.923 $\pm$ 0.018 & 0.913 $\pm$ 0.037 & 0.923 $\pm$ 0.017 & 0.918 $\pm$ 0.018 & 0.966 $\pm$ 0.015 \\
EfficientNet-B1 & 0.925 $\pm$ 0.013 & 0.910 $\pm$ 0.034 & 0.933 $\pm$ 0.034 & 0.921 $\pm$ 0.014 & 0.970 $\pm$ 0.007 \\
ResNet-18      & 0.927 $\pm$ 0.022 & 0.904 $\pm$ 0.031 & \textbf{0.943 $\pm$ 0.016} & 0.923 $\pm$ 0.022 & 0.968 $\pm$ 0.013 \\
ResNet-34      & 0.917 $\pm$ 0.018 & 0.888 $\pm$ 0.038 & 0.941 $\pm$ 0.011 & 0.913 $\pm$ 0.017 & 0.965 $\pm$ 0.006 \\
ResNet-50      & \textbf{0.930 $\pm$ 0.016} & \textbf{0.915 $\pm$ 0.046} & 0.940 $\pm$ 0.029 & \textbf{0.926 $\pm$ 0.015} & \textbf{0.970 $\pm$ 0.009} \\
\midrule
\multicolumn{6}{l}{\textbf{Traditional ML (Best F1 per Feature Family)}} \\
\midrule
SatCLIP-V + RF + Mean                 & 0.716 $\pm$ 0.017 & {0.674 $\pm$ 0.021} & \textbf{0.751 $\pm$ 0.018} & \textbf{0.710 $\pm$ 0.015} & 0.781 $\pm$ 0.011 \\
Handcrafted + XGB + PCA             & \textbf{0.718 $\pm$ 0.013} & \textbf{0.703 $\pm$ 0.014} & 0.678 $\pm$ 0.031 & 0.690 $\pm$ 0.018 & \textbf{0.786 $\pm$ 0.012} \\
GeoRSCLIP + LR + PCA                & 0.690 $\pm$ 0.022 & 0.662 $\pm$ 0.019 & 0.674 $\pm$ 0.045 & 0.668 $\pm$ 0.030 & 0.751 $\pm$ 0.019 \\
Satlas Pretrain + LR + Concat       & 0.623 $\pm$ 0.021 & 0.591 $\pm$ 0.026 & 0.610 $\pm$ 0.035 & 0.599 $\pm$ 0.022 & 0.676 $\pm$ 0.011 \\
Prithvi EO 2.0 + LR + PCA            & 0.597 $\pm$ 0.038 & 0.563 $\pm$ 0.040 & 0.570 $\pm$ 0.058 & 0.566 $\pm$ 0.048 & 0.635 $\pm$ 0.029 \\
SatMAE + GB + Concat                & 0.606 $\pm$ 0.023 & 0.577 $\pm$ 0.025 & 0.553 $\pm$ 0.033 & 0.565 $\pm$ 0.027 & 0.640 $\pm$ 0.018 \\
DINOv3 + RF + Median                & 0.596 $\pm$ 0.028 & 0.566 $\pm$ 0.031 & 0.547 $\pm$ 0.035 & 0.556 $\pm$ 0.032 & 0.621 $\pm$ 0.022 \\
\bottomrule
\end{tabular}%
}
\end{table*}

\subsection{Overall performance}
Tables~\ref{tab:cnn_and_traditional_results} summarizes performance on 2023 imagery for the CNN and traditional models. All pretrained CNNs exceed 0.91 F1; ResNet-50 achieves the best mean F1 ($0.926\pm0.015$). The best traditional ML configuration SatCLIP-V+RF+Mean) reaches F1 $0.710 \pm0.015$.
%
% \subsection{Ablations: pretraining and masking}
Table~\ref{tab:cnn_merged_comparison} shows that ImageNet pretraining improves F1 for all backbones (up to +0.143 for ResNet-34) and spatial masking is critical, yielding +0.301 to +0.455 F1 improvements by suppressing non-site context.
%
% \subsection{Traditional ML analysis: aggregation and masking effects}
 We also analyze the impact of spatial masking on traditional ML features (Table~\ref{tab:masking_ablation}); unlike CNNs, masking can degrade several foundation-model embeddings, suggesting that frozen feature extractors may depend on the context of the larger scene.
 We present in Table~\ref{tab:aggregation_comparison} the optimal configuration for each temporal aggregation approach, where tree-based classifiers (RF and XGB) consistently deliver the best performance.

% \begin{figure}[t]
% \centering
% \includegraphics[width=0.48\textwidth]{traditional_vs_cnn.png}
% \caption{Best CNN (ResNet-50) vs. best traditional ML (handcrafted + XGB + PCA) on 2023 imagery. Values are means across 5 folds; see Tables~\ref{tab:cnn_results} and~\ref{tab:best_traditional_ml_results} for exact numbers.}
% \label{fig:cnn_vs_traditional_ml}
% \end{figure}

% \begin{figure*}[t]
% \centering
% \includegraphics[width=0.95\textwidth]{detailed_comparison_traditiona_cnn_masked.png}
% \caption{Comprehensive comparison: best configuration per traditional ML feature type (blue) vs. CNN architectures (orange) across key metrics on 2023 imagery. Bars show means across 5 folds; the gap highlights the advantage of pretrained CNNs with masking.}
% \label{fig:comprehensive_comparison}
% \end{figure*}

\begin{table}[t]
\centering
\caption{F1 comparison: ImageNet pretrained vs. random initialization (with masking), and with spatial masking vs. without masking (ImageNet-pretrained). Values are mean across 5 folds; $\Delta$ is the difference. Ref.: the top-performing reference setup, i.e., the pretrained using a mask. Random refers to the random initialization of the model's weights.}
\label{tab:cnn_merged_comparison}
\small
\rowcolors{2}{gray!12}{white}
\resizebox{0.99\linewidth}{!}{%
\begin{tabular}{l c c c c c}
\toprule
  & & \multicolumn{2}{c}{\textbf{Pretrained impac}t} & \multicolumn{2}{c}{\textbf{Mask impact}} \\
  \cmidrule(r){3-4} \cmidrule(l){5-6}
\textbf{Model} & \textbf{Ref.} & \textbf{Random} & \textbf{$\Delta$ F1} & \textbf{No mask} & \textbf{$\Delta$ F1} \\
\midrule
ResNet-18 & 0.923 & 0.807 & \textbf{+0.116} & \textbf{0.613} & +0.310 \\
ResNet-34 & 0.913 & 0.770 & +0.143 & 0.564 & +0.349 \\
ResNet-50 & \textbf{0.926 }& 0.802 & +0.124 & 0.565 & +0.361 \\
EfficientNet-B0 & 0.918 & \textbf{0.862} & +0.056 & 0.463 & +\textbf{0.455} \\
EfficientNet-B1 & 0.921 & 0.844 & +0.077 & 0.620 & +0.301 \\
\bottomrule
\end{tabular}%
}
\end{table}

\begin{table}[t]
\centering
\caption{Impact of spatial masking on traditional ML performance (F1). $\Delta$F1 is masked minus unmasked; Acc = accuracy.}
\label{tab:masking_ablation}
\small
\rowcolors{2}{gray!12}{white}
\resizebox{1.0\linewidth}{!}{
\begin{tabular}{@{}lcccc@{}}
    \toprule
    \textbf{Feature Type} & \textbf{Masked F1} & \textbf{Unmasked F1} & \textbf{$\Delta$F1} & \textbf{$\Delta$Acc} \\
\midrule
SatCLIP-V & \textbf{0.710} & 0.673 & \textbf{+0.037}  & +\textbf{0.041} \\
Handcrafted & 0.690 & 0.659 & {+0.031} & +0.036 \\
Satlas Pretrain & 0.599 & 0.571 & {+0.028} & +0.022 \\
GeoRSCLIP & 0.668 & \textbf{0.676} & -0.008 & -0.012 \\
DINOv3 & 0.556 & 0.588 & -0.032 & -0.025 \\
SatMAE & 0.565 & 0.615 & -0.050 & -0.035 \\
Prithvi EO 2.0 & 0.566 & 0.623 & -0.057 & -0.037 \\
\bottomrule
\end{tabular}
}
\end{table}

\begin{table}[t]
\centering
\caption{Best configuration per temporal aggregation method (2023 imagery). Values are mean (std) across 5 folds. Best Config:  Combination of a feature set and a traditional ML model that achieves the highest F1 score for each aggregation type: HF: Handcrafted, XGB = XGBoost, LR = Logistic Regression, GB = Gradient Boosting.}
\label{tab:aggregation_comparison}
\small
\rowcolors{2}{gray!12}{white}
\resizebox{1.0\linewidth}{!}{
\begin{tabular}{@{}llcc@{}}
    \toprule
    \textbf{Aggregation} & \textbf{Best Config} & \textbf{F1} &  \textbf{AUROC} \\
\midrule
PCA & HF + XGB & $0.690 \pm0.018$ &  $0.786 \pm0.012$ \\
% Concat & HF + LR & 0.678 \pm0.013 & 0.690 \pm0.011 & 0.760 \pm0.008 \\
% Mean & HF + LR & 0.641 \pm0.017 & 0.638 \pm0.023 & 0.705 \pm0.015 \\
% Median &HF + LR & 0.635 \pm0.014 & 0.633 \pm0.020 & 0.699 \pm0.014 \\ \midrule
Concat & SatCLIP-V + XGB & $0.704 \pm 0.019$ & \textbf{0.788 $\pm$ 0.006} \\
Mean & SatCLIP-V + RF & \textbf{0.710 $\pm$ 0.015} & $0.781 \pm 0.011$ \\
Median & SatCLIP-V + RF & $0.709 \pm0.015$  & $0.781 \pm0.012$ \\
\bottomrule
\end{tabular}
}
\end{table}

\subsection{Across-year performance}
\begin{figure}[t]
    \centering
    \includegraphics[width=0.495\textwidth]{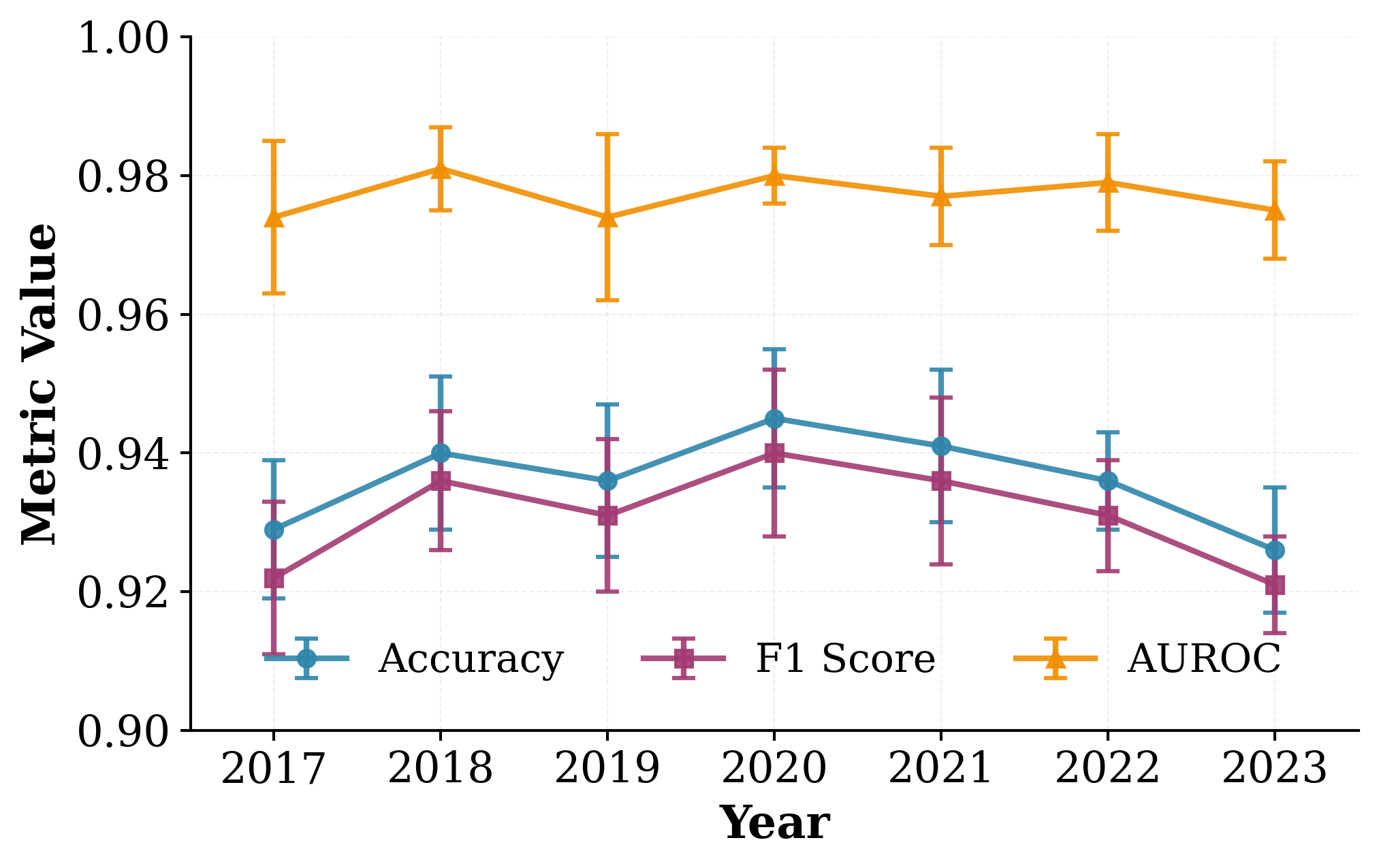}
    \caption{EfficientNet-B1 performance across individual years (2017--2023), pretrained with spatial masking. Error bars show std across folds.}
    \label{fig:temporal_year_analysis}
\end{figure}

We independently train EfficientNet-B1 on each individual year (2017–2023), using the same site splits for every fold (see Figure~\ref{fig:temporal_year_analysis}). The best performance (F1 $\approx 0.94$) is observed for 2020, indicating that most looting events annotated in the dataset occurred before 2021. The observed decline in performance indicates the presence of temporal label noise where the visible effects of earlier looting activities may fade over time.

% \begin{figure}[t]
% \centering
% \includegraphics[width=0.48\textwidth]{shap_bar_top10.png}
% \caption{Mean absolute SHAP values for top handcrafted features under an XGBoost baseline, highlighting the benefits of texture features and spectral statistics to detect looting.}
% \label{fig:shap_importance}
% \end{figure}
\subsection{Feature importance}
Table~\ref{tab:handcrafted_features} presents the top 10 handcrafted features ranked by mean absolute SHAP value, identifying the most discriminative spectral and textural signatures for looting detection. Features were selected through a two-stage process: first, ensemble feature importance was computed by averaging tree-based importances (Gini/gain) across Random Forest, Gradient Boosting, and XGBoost models trained on all 42 handcrafted features (after removing zero-variance features); the top 10 features from this ranking were then retained and used to retrain XGBoost. SHAP values were computed on held-out test data to quantify each feature's marginal contribution to individual predictions, with final rankings based on mean absolute SHAP values across all test samples. Sobel edge strength on NIR (SHAP=0.386) emerges as the most important feature, effectively capturing excavation boundaries through edge detection. Overall texture-based features provide stronger signals for looting detection. Particularly, GLCM texture measures make up most of the top-ranked features (7 out of 10), with contrast, entropy, homogeneity, and energy derived from the Red, Green, and NIR bands capturing the disruption patterns.

\begin{table*}[t]
\centering
\caption{Top handcrafted features by mean absolute SHAP value (global importance). Positive SHAP contributions indicate increased looting probability; values are aggregated over test folds.}
\label{tab:handcrafted_features}
\small
\rowcolors{2}{gray!12}{white}
\setlength{\tabcolsep}{6pt}
\begin{tabular}{@{}cll}
    \toprule
    \textbf{SHAP value} & \textbf{Feature} & \multicolumn{1}{c}{\textbf{Interpretation}} \\
\midrule
0.39 & Sobel NIR & Mean Sobel edge strength on NIR band (detects excavation boundaries/disturbance). \\
0.19 & GLCM Red Contrast & Local intensity variation in red band (higher under disturbed micro-relief). \\
0.19 & GLCM Green Entropy & Textural randomness/complexity in green band (spatial disorder). \\
0.17 & All Bands STD & Standard deviation across all four bands (R,G,B,NIR; overall spectral variability). \\
0.16 & GLCM Green Contrast & Local intensity variation in green band (excavation texture patterns). \\
0.15 & GLCM Red Homogeneity & Local similarity in red band (lower under rough excavation texture). \\
0.11 & GLCM NIR Entropy & Textural randomness in NIR band (surface complexity/disruption). \\
0.08 & All Bands Mean & Mean across all four bands (R,G,B,NIR; overall brightness/reflectance). \\
0.07 & GLCM Red Energy & Textural uniformity in red band (altered by disturbance patterns). \\
0.06 & GLCM Red Entropy & Textural randomness in red band (spatial disorder from excavation). \\
\bottomrule
\end{tabular}
\end{table*}

\section{Discussion}
Our study highlights three practical takeaways. First, transfer learning remains a strong baseline for subtle satellite cues: ImageNet features help convergence and accuracy despite spectral-domain shifts. Second, spatial priors matter: explicit footprint masking acts as an attention prior that consistently improves CNNs and the best handcrafted pipelines. Third, while foundation-model embeddings are attractive, off-the-shelf monthly embeddings perform competitively with carefully engineered handcrafted features for this problem. Texture-based subset of the handcrafted features provide stronger signal of looting activities. 
%
% From an operational standpoint, the combination of mask annotations and small CNNs offers an efficient, scalable path for large-scale monitoring at monthly cadence.

Because spatial masks may differ systematically between looted and preserved sites, models could in principle exploit mask geometry as a shortcut. In a controlled test-time intervention, we enforced a minimum test-mask area by dilating any test mask below the training median up to the median area, while leaving train/validation masks unchanged. On 2023 imagery with EfficientNet-B1 and spatial masking, this yielded mean $\pm$ std across 5 folds: Accuracy $0.842\pm0.009$, Precision $0.766\pm0.007$, Recall $0.947\pm0.024$, F1 $0.847\pm0.010$, AUROC $0.936\pm0.012$. The near-equivalence to non-enforced runs suggests decisions rely primarily on within-footprint texture/spectral cues rather than raw mask area, but it underscores that mask definitions affect inductive bias and should be stress-tested.

\paragraph{Limitations:}
This study focuses on one country (Afghanistan), and performance may change under different geology, land use, or imaging conditions. PlanetScope monthly mosaics (4.7m/pixel) can miss very small disturbances, and monthly compositing can smooth short-lived events. Therefore, images with higher resolution can reveal additional detail.
%
% \paragraph{Mask geometry sensitivity.}

\section{Conclusion and Future Work}
We introduce a scalable framework for detecting archaeological looting from medium‑resolution satellite imagery and evaluate it on the largest dataset of archaeological sites. Our study directly compares ImageNet‑pretrained CNNs against traditional classifiers using handcrafted features, composed of spectral and texture characteristics, and embeddings from geospatial foundation models.

ResNet‑50 with spatial masking achieves an F1 score of $0.926$, substantially outperforming the best output from traditional classifiers: SatCLIP-V + Random Forest + Mean aggreagation (F1 = 0.710). These results highlight the advantages of end‑to‑end learned representations when looting signatures are subtle and spatially localized. Spatial masking provides a major performance gain by focusing the model on the disturbed footprint and suppressing distracting background context, indicating that operational systems should explicitly incorporate footprint annotations or segmentation followed by lightweight CNN classifiers. Training on a single, temporally consistent year reduces label noise without degrading performance, while PCA‑based temporal aggregation preserves discriminative dynamics with minimal dimensionality. Our SHAP-based analysis of hand-made features indicates that GLCM-driven texture descriptors offer stronger signals to identify looting.

Future work involves scaling up the framework to identify looting across broader geographic areas (e.g., Syria, Sudan and Egypt), as well as investigating active‑learning and semi‑supervised methods to lessen reliance on expert annotations while still ensuring robust, large‑scale looting detection. This approach facilitates comparative analyses between individual sites, broader regions, and distinct time periods, making it possible to examine the spatial distribution and temporal development of looting activities. In doing so, it helps experts implement timely interventions.

% \section{Ethics and Societal Impact}
% Automated looting detection supports protection of cultural heritage but also carries risks. Public release of site locations can increase threat exposure; any deployment must follow strict data governance with trusted stakeholders and adhere to national regulations. 

{\small
\bibliographystyle{ieeenat_fullname}
\bibliography{references}
}

\end{document}